\pdfoutput=1

\documentclass[11pt]{article}
\usepackage{makecell,tabularx}
\usepackage[preprint]{acl}
\usepackage{times}
\usepackage{latexsym}
\usepackage{todonotes}
\usepackage{diagbox}  
\usepackage[T1]{fontenc}

\usepackage[utf8]{inputenc}
\usepackage{colortbl}
\usepackage{multirow}

\usepackage{makecell}
\usepackage{microtype}

\usepackage{inconsolata}

\usepackage{graphicx}

\title{Dialectical Behavior Therapy Approach to LLM Prompting}

\author{Oxana Vitman \\
  University of Bremen\\ 
  \texttt{vitman at uni-bremen.de} \\ \And
  Nika Amaglobeli \\
  Texas Woman's University \\ \And
  Paul Plachinda \\ 
  Portland State University\\
}

\begin{document}
\maketitle
\begin{abstract}
Large language models demonstrated state-of-the-art results on various reasoning tasks when applying the chain-of-thought (CoT) prompting technique. CoT prompting guides the model into breaking tasks into a few intermediate steps and provides step-by-step demonstrations. However, solving complex reasoning tasks remains a challenge. In this paper, we propose a novel prompting strategy inspired by Dialectical Behavioral Therapy (DBT). DBT, a form of cognitive-behavioral therapy, aims to help individuals cope with stress by developing a system of reasoning. We applied DBT's basic concepts of shaping dialog to construct prompts and conducted experiments on different datasets and LLMs with various numbers of parameters. Our results show that prompts crafted with DBT techniques significantly improve results on smaller models, achieving a 7\% increase in accuracy on the StrategyQA, 4.8\% on Aqua dataset using \textit{8b} parameters model, and a 16.2\% increase on the StrategyQA, 5.3\% on GSM8K dataset with \textit{14b} parameters model. 

   
\end{abstract}

\section{Introduction}

The recent advancements in large language models (LLMs) have demonstrated that with refined prompting, LLMs can exhibit emergent capabilities in complex understanding and question-answering tasks~\citep{chowdhery2023palm}. In particular, the introduction of chain-of-thought (CoT) prompting~\citep{wei2022chain, chowdhery2023palm} has enabled LLMs to tackle reasoning tasks such as arithmetic reasoning, commonsense reasoning, and symbolic reasoning. 

CoT prompting combines natural language rationales~\citep{ling-etal-2017-program, cobbe2021training}  with few-shot prompting~\citep{brown2020language}. Enhanced with self-consistency decoding~\citep{wang2022self} in place of the traditional greedy decoding, few-shot chain-of-thought prompting achieves state-of-the-art results on numerous challenging natural language processing tasks~\citep{brown2020language}, while remaining fully interpretable.

However, chain-of-thought prompting faces a critical limitation: it often underperforms on tasks that require conceptual reasoning, such as abstract algebra and school mathematics~\citep{rae2021scaling}.

While LLMs possess the knowledge to solve such problems in a few steps, zero-shot prompting requires careful navigation to leverage that knowledge.

To tackle these challenges, we propose a novel approach inspired by Dialectical Behavior Therapy (DBT)~\citep{Chapman2006}. DBT is a cognitive-behavioral approach aimed at improving emotional regulation and interpersonal skills through its core principles, such as Wise Mind, Observation, Description, and Effectiveness. 
In the context of therapy, this often means developing a system of reasoning that helps to find balance through the exchange of logical arguments~\cite{Swales2016}.

In this paper, we hypothesize that the prompts constructed using DBT techniques are expected to yield higher accuracy on various datasets. The foundation for this hypothesis comes from psychology, where DBT is widely used to enhance logical reasoning, the ability to generalize, and attention to detail.





\section{DBT Prompting Technique}

\begin{figure*}[hbt!]
\centering
\includegraphics[width=\textwidth]{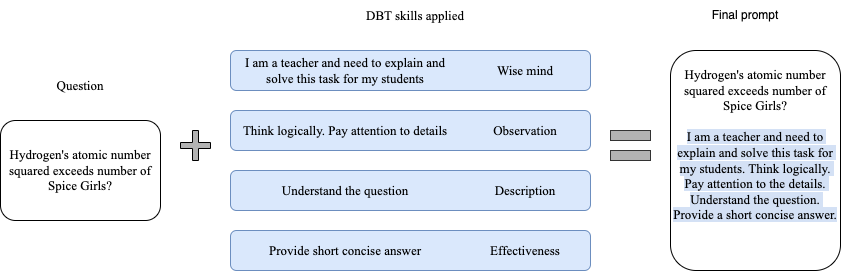}
\caption{DBT prompting technique.}
\label{fig:experiments}
\end{figure*}

To guide LLMs in generating more efficient answers to reasoning tasks, we construct prompts that combine several aspects, or ``skills,'' derived from DBT methods.
We performed a series of experiments on various datasets, where each input to the model was accompanied by several DBT suffixes, which were added after each question. These suffixes were identical across all datasets and models. Figure~\ref{fig:experiments} illustrates the design of the experiment.

\subsection{DBT Prompts}\label{sec:prompt}

This section presents two prompts we devised using DBT techniques, referred to as \textit{DBT\_1} and \textit{DBT\_2}. 

\textit{DBT\_1} = \textbf{``I am a teacher and need to explain and solve this task for my students. Think logically. Pay attention to details. Understand the question. Provide a short, concise answer.''}. This prompt is constructed based on the following DBT skills:
\begin{itemize}
    \item Wise Mind: Implied in the entire prompt, it underlies the overall approach of using both logical reasoning and intuitive understanding to explain and solve the task effectively.
\item Observation: Explicitly highlighted in ``Pay attention to details'', which directs the focus on observing and processing key aspects of the task and the student's needs.
\item Description: Reflected in ``Understand the question'', which emphasizes the importance of clearly grasping and articulating the problem.
\item Effectiveness: Captured in ``Provide a short, concise answer'', which aims to deliver a practical and efficient solution to the task. 
\end{itemize}

\textit{DBT\_2} = \textbf{``Integrate logical analysis and intuitive understanding for each question. Pay meticulous attention to every detail in the questions and choices. Clearly identify and articulate the main point of each question. Approach each question without bias or assumptions. Select the most accurate answer efficiently and confidently. Provide answers in a single-letter format (A, B, C, or D).''}. DBT skills applied:
\begin{itemize}
    \item Wise Mind: Reflected in ``Integrate logical analysis and intuitive understanding for each question'', combining rational reasoning with intuitive insights.
\item Observation: Evident in ``Pay meticulous attention to every detail in the questions and choices'', emphasizing careful and detailed examination.
\item Description: Present in ``Clearly identify and articulate the main point of each question,'' focusing on precise and accurate problem representation.
\item Effectiveness: Seen in ``Select the most accurate answer efficiently and confidently'', aiming for optimal outcomes with a streamlined approach.

\end{itemize}
The last sentence (Effectiveness) changes depending on whether the dataset's ground truth answer is a yes/no, multiple choice, or numeric.

\section{Experimental Setup}

\subsection{Datasets}
\label{sec:ds}
We evaluated the DBT-based prompting technique on datasets of various reasoning types.
\begin{itemize}
\item \textbf{Arithmetic reasoning}. For this task, we used the math word problems dataset SVAMP~\citep{patel-etal-2021-nlp}, algebraic word problems AQUA~\citep{ling-etal-2017-program}, high-quality math problems GSM8K~\citep{cobbe2021training}.
\item \textbf{Commonsense reasoning}. We used the yes/no creative questions from the StrategyQA dataset~\citep{Geva2021DidAU}.

\end{itemize}
Selected datasets feature definite numerical, yes/no, or multiple-choice answers, which dramatically simplify parsing and answer assessment.

\subsection{Baselines}

\subsubsection{Zero-shot}

We refer to zero-shot prompting as simply passing the task to the model without any additions and parsing the LLM's response.

\subsubsection{CoT}

There are two main approaches to CoT prompting. The first approach, termed Zero-Shot CoT, involves adding a simple prompt, ``Let's think step by step'' after a question, without specific examples or training on particular tasks~\citep{kojima2022large}. This method enables LLMs to perform adequately as zero-shot reasoners. The second approach entails providing a series of manually crafted reasoning demonstrations combined with few-shot prompting~\citep{wei2022chain}. Each demonstration consists of a question, a response, and a reasoning chain that includes multiple intermediate steps. Since these demonstrations are crafted manually, this approach is referred to as Manual-CoT. In our experiments, we compare the DBT approach with both CoT methods.

\subsubsection{Plan and solve}
Plan and solve~\citep{wang2023plan} (PS) prompting is a zero-shot prompting method that enables LLMs to plan how to solve a problem and generate the intermediate reasoning steps before arriving at the final answer. Unlike the Manual-CoT method, PS prompting does not incorporate step-by-step demonstration examples. It requires only the problem itself and a trigger sentence in the prompt. While the CoT method suggests \textit{``Let's think step by step''} as a trigger sentence, PS simply replaces it with \textit{``Let’s first understand the problem and devise a plan to solve the problem. Then, let’s carry out the plan and solve the problem step by
step''}.


\newcolumntype{Y}{>{\centering\arraybackslash}X}

\begin{table*}[!ht]
\caption{Accuracy score for various prompt engineering methods on different datasets. Three models are used: llama3 \cite{meta2024introducing}, phi3:medium \cite{microsoft2023phi3medium}, GPT-3.5-turbo Instruct \cite{openai2021gpt35turbo}. }
\label{table:results}
\centering
\begin{tabularx}{1\textwidth}{@{}lYYYYY@{}}
\hline
 & \multicolumn{5}{c}{Datasets} \\
\cline{2-6}
Method & Svamp & Aqua & GSM8K & StrategyQA  & Averaged accuracy \\
\hline

\multicolumn{6}{c}{\textbf{\textit{llama3}}} \\
\hline
Zero-shot & 77.6 & 44.0 & \textbf{64.0} & 24.4 & 52.5 \\
CoT & 76.2 & 39.0 & 63.9 & 21.0 & 50.0 \\
CoT + demo & 78.1 & 35.8 & 62.2 & 48.1 & 56.0 \\
Plan and Solve & 74.7 & 31.5 & 56.0 & 33.0 & 48.8 \\
\textit{DBT\_1} & 78.6 & 42.1 & 60.6 & 31.0 & 53.1 \\
\textit{DBT\_1 + demo} & \textbf{78.9} & 35.0 & 63.0 & \textbf{55.1} & \textbf{58.0} \\
\textit{DBT\_2} & 78.0 & \textbf{48.8} & 52.0 & 49.0 & \textbf{57.0} \\
\textit{DBT\_2 + demo} & 55.0 & 29.1 & 31.3 & \textbf{54.8} & 42.6 \\
\hline

\multicolumn{6}{c}{\textbf{\textit{phi3:14b}}} \\
\hline
Zero-shot & 75.1 & 52.4 & 65.2 & 35.8 & 57.1 \\
CoT & \textbf{78.2} & 54.0 & 68.3 & 24.2 & 56.1 \\
CoT + demo & 75.3 & \textbf{57.9} & 68.6 & 42.9 & 61.1 \\
Plan and Solve & 73.0 & 53.7 & 65.0 & 27.0 & 54.7 \\
\textit{DBT\_1} & 75.0 & 55.2 & 68.0 & 34.0 & 58.1 \\
\textit{DBT\_1 + demo} & 75.0 & 53.2 & 67.4 & 49.0 & 61.2 \\
\textit{DBT\_2} & 66.0 & 56.0 & \textbf{74.0} & \textbf{54.0} & \textbf{62.5} \\
\textit{DBT\_2 + demo} & 74.0 & 49.2 & \textbf{71.0} & \textbf{59.0} & \textbf{63.3}\\
\hline

\multicolumn{6}{c}{\textbf{\textit{GPT-3.5-turbo Instruct}}} \\
\hline
Zero-shot & 63.4 & 21.0 & 51.0 & 39.0 & 43.6 \\
CoT & 73.3 & 52.0 & 63.4 &\textbf{67.3} & 64.0 \\
CoT + demo & \textbf{82.2} & 52.5 & \textbf{66.3} & 59.4 & \textbf{65.1} \\
Plan and Solve & 61.4 & \textbf{62.0} & 53.5 & 56.0 & 58.2 \\
\textit{DBT\_1} & 60.4 & 36.6 & 55.4 & 37.6 & 37.1 \\
\textit{DBT\_1 + demo} & 63.4 & 39.6 & 46.5 & 53.5 & 46.6 \\
\textit{DBT\_2} & 61.2 & 37.4 & 51.3 & 58.9 & 52.2 \\
\textit{DBT\_2 + demo} & 66.3 & 27.7 & 47.5 & 55.5 & 49.3 \\
\hline
\end{tabularx}
\end{table*}

\section{Results}
We conducted a series of experiments applying the DBT-based prompting techniques on various datasets, using three LLMs with different numbers of parameters, and compared the performance of our method with baselines. 

We experimented with DBT prompts in combination with demonstrations, similar to CoT prompting, and without them. DBT prompts with demonstrations represented in Table~\ref{table:results} as \textit{DBT\_1 + demo} or \textit{DBT\_2 + demo}.

Table~\ref{table:results} shows that our method achieves best results and outperforms baselines on the smaller \textit{llama3} model with \textit{8b} parameters~\citep{meta2024introducing}. Particularly, in comparison to baselines, the accuracy increased by 4.8\% on the Aqua dataset, by 7\% on the StrategyQA dataset using prompt without demonstrations, and by 3\% on the StrategyQA dataset. 

Performance on the larger model \textit{phi3:medium} with \textit{14b} parameters~\citep{microsoft2023phi3medium} excels over baselines on two datasets: in comparison to CoT baseline, the accuracy of our method is higher on GSM8K dataset by 5.4\% (prompting without demonstrations) and on StrategyQA by 16.2\%. 
Finally, the results on the largest model, GPT-3.5-turbo Instruct~\citep{openai2021gpt35turbo}, do not exceed the baselines.

Interestingly, the largest model generates better answers when demonstrations are provided, although this is not necessarily the case for smaller models. 
On the contrary, DBT prompts designed without demonstrations either exceed or perform on par with those where the demo has been provided. This particularly contrasts with CoT, where utilization of the demo is a crucial component to increase accuracy. 


However, including demonstrations into the prompt can limit the model’s flexibility when faced with novel or unexpected problems, as its performance heavily relies on the similarity of new issues to the provided examples~\citep{brown2020language}. In contrast, our prompts bypass this reliance on demonstrations, allowing the model to broadly leverage its general problem-solving abilities and apply its reasoning skills. 


As a demonstration of the advantage of DBT-based prompting on smaller models, we present a graph with average accuracy over each method across all datasets within one model (see Appendix~\ref{sec:appendix}).

\section{Conclusion and Future Studies}

We demonstrated that the DBT prompt construction technique significantly enhances response accuracy in more compact LLMs, such as \textit{llama} and \textit{phi3}. In contrast, larger models like GPT typically start with high accuracy, and additional DBT ``wrappers'' lead model to produce more responses that stray away from the main point. 

Applying DBT techniques for prompting can have many variations, as DBT methods can be implemented in different ways. In that regard, we encourage future researchers to explore and develop their own interpretations of this approach.  

An important application of our DBT-based prompting extends beyond LLMs, as it is rooted in the observed similarity of response between LLMs and humans~\citep{kawakita2023gromov}. This similarity allows for the preliminary testing of DBT techniques on machines, thus facilitating the exploration of these methods for educational and psychological applications before directly involving human subjects.



\bibliography{acl_latex}

\appendix

\section{Average accuracy values for various prompts.}
\label{sec:appendix}

\begin{figure*}[t]
    \centering
    \includegraphics[width=\textwidth]{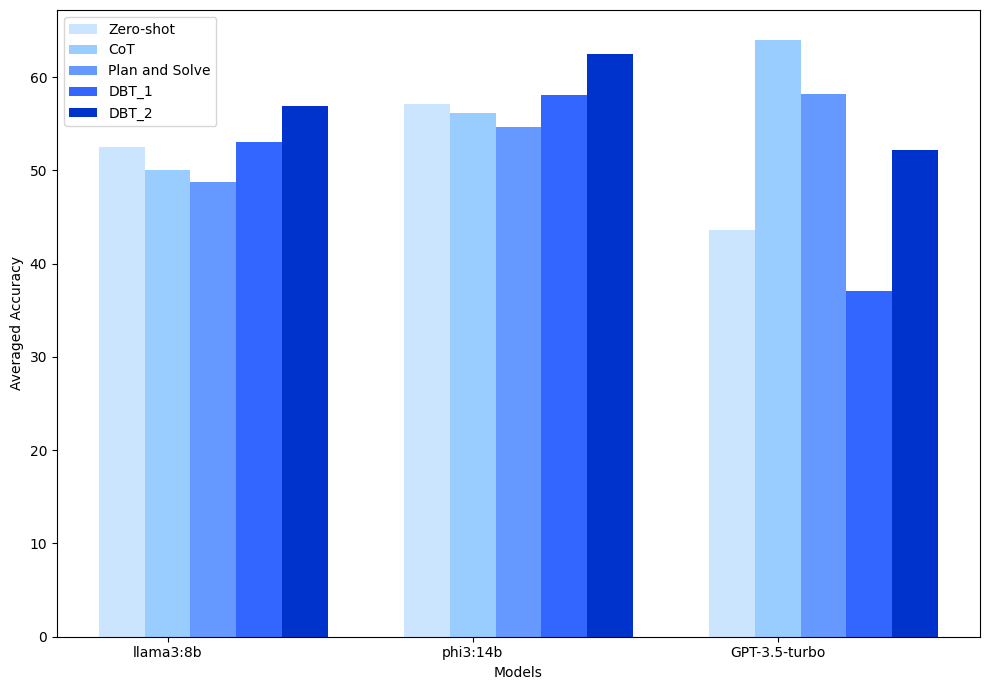}
    \caption{Average accuracy values for various prompts.}
    \label{fig:results}
\end{figure*}

We argue that this metric is important for those cases when there is a random question, and we don't have \textit{a priori} knowledge about which reasoning task the question might belong to. In Fig.\ref{fig:results}, we can see that smaller models with DBT prompting applied show better-averaged results across all datasets.

\end{document}